\documentclass[conference]{IEEEtran}
\IEEEoverridecommandlockouts
\usepackage{cite}
\usepackage{amsmath,amssymb,amsfonts}
\usepackage{algorithmic}
\usepackage{graphicx}
\usepackage{textcomp}
\usepackage{xcolor}

\usepackage{makecell}
\usepackage{multirow}
\usepackage{microtype}
\usepackage[all]{nowidow}

\def\BibTeX{{\rm B\kern-.05em{\sc i\kern-.025em b}\kern-.08em
    T\kern-.1667em\lower.7ex\hbox{E}\kern-.125emX}}
\begin{document}

\title{Learning the Pedestrian-Vehicle Interaction for Pedestrian Trajectory Prediction
\thanks{
This research is funded by the European research project ``SHAPE-IT – Supporting the Interaction of Humans and Automated Vehicles: Preparing for the Environment of Tomorrow’’. This project has received funding from the European Union’s Horizon 2020 research and innovation programme under the Marie Skłodowska-Curie grant agreement 860410.
}
}

\author{\IEEEauthorblockN{Chi Zhang}
\IEEEauthorblockA{\textit{Department of Computer Science and Engineering} \\
\textit{University of Gothenburg}\\
Gothenburg, Sweden \\
chi.zhang@gu.se
}
\and
\IEEEauthorblockN{Christian Berger}
\IEEEauthorblockA{\textit{Department of Computer Science and Engineering} \\
\textit{University of Gothenburg}\\
Gothenburg, Sweden \\
christian.berger@gu.se
}
}

\maketitle
\begin{abstract}
In this paper, we study the interaction between pedestrians and vehicles and propose a novel neural network structure called the Pedestrian-Vehicle Interaction (PVI) extractor for learning the pedestrian-vehicle interaction. We implement the proposed PVI extractor on both sequential approaches (long short-term memory (LSTM) models) and non-sequential approaches (convolutional models). We use the Waymo Open Dataset that contains real-world urban traffic scenes with both pedestrian and vehicle annotations. For the LSTM-based models, our proposed model is compared with Social-LSTM and Social-GAN, and using our proposed PVI extractor reduces the average displacement error (ADE) and the final displacement error (FDE) by 7.46\% and 5.24\%, respectively. For the convolutional-based models, our proposed model is compared with Social-STGCNN and Social-IWSTCNN, and using our proposed PVI extractor reduces the ADE and FDE by 2.10\% and 1.27\%, respectively. The results show that the pedestrian-vehicle interaction influences pedestrian behavior, and the models using the proposed PVI extractor can capture the interaction between pedestrians and vehicles, and thereby outperform the compared methods.

\end{abstract}


\section{Introduction}
The prediction of a pedestrian's future trajectory is crucial for socially-aware robots and automated vehicles (AVs) to prevent collisions and other hazardous situations.
In urban traffic scenarios, the prediction of pedestrians' trajectories is a great challenge, because pedestrians can change their direction and velocity suddenly~\cite{volz2016data,volz2015feature}. Besides, the movement of pedestrians depends on their own past trajectories and is influenced by social interactions with others~\cite{moussaid2010walking}. Furthermore, pedestrians are influenced by surroundings~\cite{volz2015feature} and other road users such as vehicles.

Much work has been devoted to predicting pedestrians' trajectories using deep learning methods, including sequential models such as long short-term memory (LSTM) networks~\cite{alahi2016social}, LSTM-based generative adversarial networks (GANs)~\cite{gupta2018social} and non-sequential models such as convolutional neural networks (CNNs)~\cite{mohamed2020social, nikhil2018convolutional}.
However, to the best knowledge of the authors, one of the most significant influences for pedestrian behavior originating by multiple surrounding vehicles is not considered by most works so far. Recent work by Eiffert et al.~\cite{eiffert2020probabilistic} tried to include such influences but considered only a single vehicle around pedestrians. Some recent works predict the trajectories of heterogeneous traffic agents including pedestrians but usually focus more on vehicles or motorcycles~\cite{chandra2019traphic,Chandra2019Robusttp,Chandra2020,Carrasco2021scout}.
Besides, the ETH~\cite{pellegrini2009you} and UCY~\cite{lerner2007crowds} datasets, which are commonly used to evaluate pedestrian trajectory predictions, do not include annotations for vehicles, and hence, are not suitable for an appropriate evaluation of this task. 

Therefore, we focus on the prediction of pedestrian trajectories, considering the influence of pedestrian-vehicle interactions.
The main contributions of this paper are as follows:
\begin{itemize}
    \item We have proposed the Pedestrian-Vehicle Interaction (PVI) extractor to predict pedestrian trajectories. The features of interactions between pedestrians and vehicles are encoded by the vehicle feature embedding layers and pedestrian-vehicle interaction module.
    
    \item We have implemented, evaluated, and analyzed the proposed PVI extractor on both sequential (LSTM-based) and non-sequential (convolutional-based (Conv-based)) models. The LSTM-based model using our proposed PVI extractor is compared against Social-LSTM~\cite{alahi2016social} and Social-GAN~\cite{gupta2018social}, and reduces the average displacement error (ADE) and the final displacement error (FDE) by 7.46\% and 5.24\%, respectively, compared to Social-LSTM. The Conv-based model using our proposed PVI extractor is compared against Social-STGCNN~\cite{mohamed2020social} and Social-IWSTCNN~\cite{zhang2021social}, and outperforms Social-STGCNN on ADE and FDE by 2.10\% and 1.27\%, respectively. The results show the efficiency of the proposed PVI extractor.
    
    \item As we aim to solve the real-world task of forecasting the trajectories in urban traffic scenarios, the proposed algorithm is trained and evaluated on real-world urban traffic data using the Waymo Open Dataset~\cite{sun2020scalability}.
\end{itemize}

\section{Related Work}
\label{sec:RelatedWork}
\subsection{Pedestrian Trajectory Prediction}
\label{sec:pedestrian_trajectory_prediction}

\paragraph{Sequential Models}
Recent research has demonstrated the potential of recurrent networks on pedestrian trajectory prediction.
Alahi et al.~\cite{alahi2016social} proposed Social-LSTM to predict pedestrian trajectories that applied a social pooling layer over LSTMs to represent the interaction between pedestrians instead of using hand-crafted models like Social-Force ~\cite{helbing1995social}.
Some later works extended Social-LSTM by improving the interaction aggregation module or by including the environment or neighbors feature extracted from 2D images.
Xue et al.~\cite{xue2018ss} introduced scene information to the LSTM-based framework, and proposed Social-Scene-LSTM, which used two additional LSTMs to handle the neighboring pedestrians and environment information. A combined attention model is applied over LSTM by Fernando et al.~\cite{fernando2018soft+}, which utilizes both ``soft attention'' and ``hard-wired attention'' in order to map the trajectory information from the local neighborhood to future positions. Traphic~\cite{chandra2019traphic} used LSTMs with the appearance feature extracted from neighbors and horizon view images using CNNs.

Unlike the uni-modal distribution assumption made by previous LSTM-based models, Gupta et al.~\cite{gupta2018social} argued that multiple trajectories are plausible given the history of a trajectory. The authors assumed that the pedestrian trajectories follow a multi-modal distribution, and proposed Social-GAN with LSTM-based generators.
Following this assumption, Social-ways~\cite{amirian2019social} improves the attention pooling structure and utilizes the info-GAN without L2 loss. SoPhie~\cite{sadeghian2019sophie} considered the environment context feature extracted from 2D images by CNNs within the LSTM-based GAN framework and improved the social interaction attention mechanism. CGNS~\cite{li2019conditional} used gate recurrent units (GRUs) as the encoder and used the environment information to get better performance based on conditional variational autoencoders (CVAEs) models. Social-BiGAT~\cite{kosaraju2019social} was also based on GANs but instead of using the pooling module, it utilizes the graph attention (GAT) network over the hidden states of LSTMs to learn the social attention of pedestrians.

\paragraph{Non-sequential Models}
Nikhil and Morris~\cite{nikhil2018convolutional} revealed that CNNs can be used for trajectory prediction and can even achieve competitive results while reaching computational efficiency. Unlike RNNs, whose later time-steps prediction depends on previously predicted time-steps, Conv-based methods predict all time-steps at once, and hence, it can reduce the accumulated error. Bai et al.~\cite{bai2018empirical} pointed out that the usage of Temporal Convolutional Networks (TCNs) can be used for sequence modeling instead of RNNs because of their efficiency. With the help of TCNs and CNNs, Social-STGCNN~\cite{mohamed2020social} made a breakthrough on improving the accuracy with a faster speed. Zhang et al.~\cite{zhang2021social} followed this trend of using TCNs and CNNs for predicting, and they proposed a sub-network to learn the social attention weights to improve the results and removed the graph construction to save inference time.

\subsection{Modeling the Interaction Between Pedestrians and Vehicles}
Most existing works considered the social interaction with other pedestrians~\cite{alahi2016social,gupta2018social,sadeghian2019sophie,mohamed2020social, kosaraju2019social} and the interaction with the environment~\cite{Manh2018,xue2018ss,sadeghian2019sophie}.

However, the interaction between pedestrians and vehicles, which is an equally important factor, has not been investigated much in previous research. For urban traffic scenarios, some researchers tried to include vehicle information to get more precise prediction results. Eiffert et al.~\cite{eiffert2020probabilistic} improved pedestrian trajectory prediction by a feature learning network, that encodes interactions between pedestrians and a single vehicle by the ``Graph pedestrian-vehicle Attention Network''. However, they only included one single car but not multiple vehicles on the road.
Chandra et al.~\cite{chandra2019traphic,Chandra2019Robusttp,Chandra2020} and Carrasco et al.~\cite{Carrasco2021scout} proposed several models that predict the trajectories of heterogeneous traffic agents including pedestrians, but they focused primarily on vehicles and motorcycles rather than pedestrians.

In this paper, we focus on pedestrian prediction and consider the influence of \emph{all} vehicles in a frame. We propose the Pedestrian-Vehicle Interaction (PVI) extractor to extract the interactions between \emph{all vehicles and target pedestrians}, and apply the module on both sequential and non-sequential prediction models.
We apply the proposed PVI extractor to both sequential models (LSTMs) and non-sequential models (Conv-based models including CNNs and TCNs) to show the effectiveness of our algorithm. For the LSTM-based model, we extend the Social-LSTM model by introducing the pedestrian-vehicle interaction feature extracted with the proposed PVI extractor. Meanwhile, we propose a more effective structure for extracting social interaction features compared with Social-LSTM.
For the Conv-based model, we extend our previous work in Social-IWSTCNN~\cite{zhang2021social} by introducing the pedestrian-vehicle interaction with the proposed PVI extractor in an additional feature extracting stream. Besides, we investigate the effect of different inputs for learning the interaction relationship, including relative position and relative velocity.

\section{Methodology}
\label{sec:Methodology}

\subsection{Problem Definition}
\label{ProblemDefination}
Given observed trajectories of pedestrians and vehicles in a sequence, we aim to predict the most likely trajectories of pedestrians in the future. The positions of pedestrians and vehicles in each frame are first pre-processed to $x$ and $y$ coordinates on a 2D map representation in bird's-eye-view. 
In a frame at any time-step $t$ with the number of pedestrians $n_p$ and the number of vehicles $n_v$, the observation of pedestrians and vehicles can be denoted as $X_t = [X_t^1, X_t^2, \dots, X_t^{n_p}] $, $V_t = [V_t^1, V_t^2, \dots, V_t^{n_v}]$, with all observed time-steps $1 \leq t \leq T_{obs}$.
The input x-y-coordinates of the $i^{th}$ person at time $t$ is defined as $X_t^i = (x_t^i, y_t^i)$, where $i\in{\{1,\dots, n_p\}}$, and the input x-y-coordinates of the $j^{th}$ vehicle at time $t$ is defined as $V_t^j = (x_t^j, y_t^j)$, where $j \in{\{1,\dots, n_v\}}$.
The predicted trajectory of pedestrians is $\hat Y_t = [\hat Y_t^1, \hat Y_t^2, \dots, \hat Y_t^{n_p}]$, where $T_{obs}+1 \leq t \leq T_{pred}$. The ground truth of the future trajectory is denoted as $Y$, where $Y_t = [Y_t^1, Y_t^2, \dots, Y_t^{n_p}]$. 

\subsection{Overall Framework of Pedestrian Trajectory Prediction} 
\label{subsec:overall framework}
\begin{figure}
\begin{center}
\includegraphics[scale=0.35]{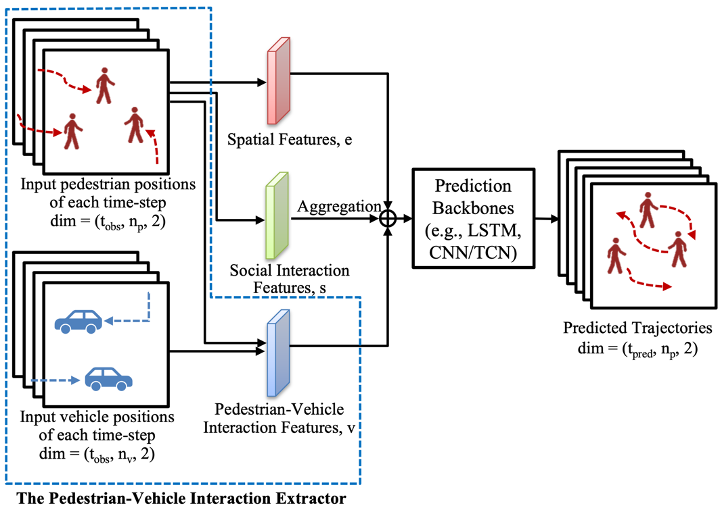}
\end{center}
  \caption{The overall framework of pedestrian trajectory prediction when considering the vehicles. The part inside the blue dotted polygon box is the proposed Pedestrian-Vehicle Interaction Extractor.}
\label{fig:overall}
\end{figure}

The overall framework for pedestrian trajectory prediction is presented in Fig.~\ref{fig:overall}. The inputs of the network include the trajectories of both pedestrians and vehicles. Three kinds of features are considered: the input spatial embedding feature ($e_t^i$), denoted as the red block, the social interaction feature between pedestrians ($s_t^i$), denoted as the green block, and the interaction feature between vehicles and pedestrians ($v_t^i$), denoted as the blue block. These features are aggregated together and followed by the prediction backbones, such as LSTMs and CNNs. 

\paragraph{Spatial Embedding Features} The spatial feature of a pedestrian is captured by embedding their input x-y-coordinate positions, as shown in Eq.~\ref{eq_input_emb}.
\begin{equation}
e_t^i = \phi(\Delta X_t^i; W_{e})
\label{eq_input_emb}
\end{equation}
where $\phi(\cdot)$ denotes the embedding function with ReLU non-linearity. $W_{e}$ is the embedded weights and $\Delta X_t^i = (x_t^i-x_{t-1}^i,y_t^i-y_{t-1}^i)$ is the relative position of the $i^{th}$ pedestrian between current time-step $t$ and the last time-step $t-1$.

\paragraph{Social Interaction Feature}
We learn the social interactions from relative positions between pedestrians and the others with a multi-layer perceptron (MLP) sub-network, as denoted in Eq.~\ref{eq_social}.
\begin{equation}
s_t^{i} = Pooling\{MLP(\phi(d_t^{ik};W_{r}); W_{s})\}, i \neq k
\label{eq_social}
\end{equation}
where $s_t^{i}$ is the social interaction feature of $i^{th}$ pedestrian at time-step $t$, where $i\in{\{1,\dots, n_p\}}$. The $MLP (\cdot)$ represents the MLP sub-network. The $d_t^{ik} = (\Delta x^{ik}, \Delta y^{ik})=(x^k - x^i, y^k - y^i)$ are the relative positions between $i^{th}$ and $k^{th}$ pedestrians, where $k\in{\{1,\dots, n_p\}}, i \neq k $. $W_r$ are linear embedding weights. $W_s$ are learned parameters of the MLP. Max pooling is used to aggregate the social interaction.

\paragraph{Pedestrian-Vehicle Interaction Feature}
We extract the pedestrian-vehicle interaction feature $v_t^i$ using PVI extractor, $PVI(\cdot)$, as denoted in Eq.~\ref{eq_veh}.
\begin{equation}
v_t^i = PVI(X_t^i, V_t^j, \Delta V_t^j), i\in{\{1,\dots, n_p\}}, j \in{\{1,\dots, n_v\}}
\label{eq_veh}
\end{equation}
where $X_t^i = (x_t^i, y_t^i)$ is the position of the $i^{th}$ pedestrian at time-step $t$; $V_t^j = (x_t^j, y_t^j)$ is the position of the $j^{th}$ vehicle; $\Delta V_t^j = (x_t^j-x_{t-1}^j,y_t^j-y_{t-1}^j)$ is the relative position of the $j^{th}$ vehicle between current time-step $t$ and last time-step $t-1$. The details of the PVI extractor are presented in Sec.~\ref{subsec:VehiclePedInteraction}.

\paragraph{Aggregation Module}
The aforementioned features are aggregated as shown in Eq.~\ref{eq_concat}.  
\begin{equation}
l_t^i = AGG(e_t^i, s_t^i, v_t^i)
\label{eq_concat}
\end{equation}
where $AGG(\cdot)$ denotes the aggregation module that aggregates the pedestrian spatial features, the social interaction features, and the pedestrian-vehicle interaction features.

The aggregation modules in the sequential model and non-sequential model are slightly different.
For the sequential model, we follow the setting in Social-LSTM~\cite{alahi2016social} and use concatenation as the aggregation module. We only calculate the interaction aggregation once at the last time-step of observation $T_{obs}$ to save inference time as stated in Social-GAN~\cite{gupta2018social}, instead of calculating the interaction at each prediction time-step as used in Social-LSTM~\cite{alahi2016social}.
For the non-sequential model, we follow the setting in Social-IWSTCNN~\cite{zhang2021social}, and use weighted sum function as aggregation module for pedestrian spatial features and social features, and then concatenate the pedestrian spatial and social feature with the pedestrian-vehicle interaction features for prediction.

\paragraph{Prediction Backbones}
Finally, the extracted features are fed into the prediction backbones $PRED(\cdot)$ to get the final prediction $o_t^i$ as shown in Eq.~\ref{eq_pred}:
\begin{equation}
o_t^i = PRED(l_t^i)
\label{eq_pred}
\end{equation}

We apply the proposed PVI extractor to two different prediction backbones including the sequential and non-sequential models.
For the sequential model, we use the LSTM encoder-decoder structure as the prediction module.
For the non-sequential models, we use Conv-based models and adopt the settings of TCNs and CNNs following the Social-STGCNN~\cite{mohamed2020social}.
By applying TCNs, we capture the temporal relationship in time-scales for sequential predictions. After extracting the temporal features, the CNN extrapolator is used to predict all time-steps of the future prediction horizon.

\subsection{Pedestrian-Vehicle Interaction (PVI) Extractor}
\label{subsec:VehiclePedInteraction}
\begin{figure}
\begin{center}
\includegraphics[scale=0.36]{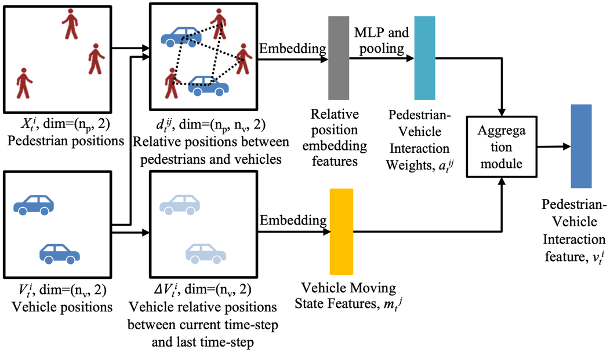}
\end{center}
  \caption{The Pedestrian-Vehicle Interaction Module.}
\label{fig:interaction}
\end{figure}

In urban traffic scenarios, pedestrians adjust their paths implicitly depending on the movement states of vehicles. Therefore, we consider the pedestrian-vehicle interaction as an influencing factor when predicting a pedestrian's trajectory.
The PVI extractor includes two streams as presented in Fig.~\ref{fig:interaction}.
We learn the pedestrian-vehicle interaction weights from the relative positions between pedestrians and vehicles as represented in the top stream.
A separate embedding module is used to extract the movement state features of the vehicles as represented in the bottom stream. Finally, we use the aggregation module to get the pedestrian-vehicle interaction features that are used for the prediction as represented by the blue block. The details of the $PVI(\cdot)$ are presented as follows.

Firstly, we learn the pedestrian-vehicle interaction attention weights $a_t^{ij}$ between the $i^{th}$ pedestrian and the $j^{th}$ vehicle as shown in Eq.~\ref{eq_vehicle_att}. Max pooling is used in this paper.
\begin{equation}
{a}_t^{ij} = Pooling \{MLP (\phi (d_t^{ij}; W_d); W_a)\},
\label{eq_vehicle_att}
\end{equation}
$$i\in{\{1,\dots, n_p\}}, j \in{\{1,\dots, n_v\}}$$
where $Pooling(\cdot)$ is the pooling layer; $MLP(\cdot)$ is the MLP layer with weights $W_a$, and $\phi (\cdot)$ is the embedding layer with weights $W_d$. The relative position between the $i^{th}$ pedestrian $X_t^i$, and the $j^{th}$ vehicle $V_t^{j}$ at time-step $t$ is denoted as $d_t^{ij}$.

With the $i^{th}$ pedestrian position $X_t^i = (x_{ped,t}^{i}, y_{ped,t}^{i})$, where $ped$ denotes pedestrians, and the $j^{th}$ vehicle $V_t^{j} = (x_{veh,t}^{j}, y_{veh,t}^{j})$, where $veh$ denotes the vehicles, the relative position $d_t^{ij}$ is calculated as shown in Eq.~\ref{eq_vehicle_rel}. All pedestrians and all vehicles in the frame are considered.
\begin{equation}
d_t^{ij}=(x_{veh,t}^{j} - x_{ped,t}^{i}, y_{veh,t}^{j} - y_{ped,t}^{i}), 
\label{eq_vehicle_rel}
\end{equation}
$$ i\in{\{1,\dots, n_p\}}, j \in{\{1,\dots, n_v\}} $$

Then, we calculate the movement state of the $j^{th}$ vehicle, denoted as $m_t^j$, as shown in Eq.~\ref{eq_vehicle_states}.
\begin{equation}
m_t^j = \phi(\Delta V_t^j; W_{m}), j \in{\{1,\dots, n_v\}}
\label{eq_vehicle_states}
\end{equation}
where $\Delta V_t^j$ is the relative position of the $j^{th}$ vehicle between current time-step $t$ and last time-step $t-1$; $\phi (\cdot)$ is the embedding layer with weights $W_m$. 

Finally, we aggregate the pedestrian-vehicle interaction weights and the vehicle movement states to get the pedestrian-vehicle interaction features.
\begin{equation}
v_t^i = AGG_{PVI}(m_t^j, a_t^{ij}),
\label{eq_vehicle}
\end{equation}
$$ i\in{\{1,\dots, n_p\}}, j \in{\{1,\dots, n_v\}} $$
where $AGG_{PVI}(\cdot)$ is the aggregation module used in the PVI extractor.
For the LSTM-based model, we follow the aggregation method in Social-LSTM~\cite{alahi2016social}. We add an embedding layer after the vehicle movement state $m_t^j$, and then concatenate the embedded feature with the pedestrian-vehicle interaction features.
For the Conv-based model, we follow and extend the aggregation method in Social-IWSTCNN~\cite{zhang2021social}, using weighted sum function to aggregate the vehicle movement state features and pedestrian-vehicle interaction weights.

\section{Experiments}
\label{sec:Experiments}
\subsection{Dataset Introduction and Data Pre-processing}
\label{sec: dataset}
Datasets are essential for the training and testing of deep learning models. ETH~\cite{pellegrini2009you} and UCY~\cite{lerner2007crowds} datasets are popular and used by many researchers to evaluate the prediction of pedestrian trajectories. However, these two datasets are not collected for urban traffic scenarios. Waymo released a real-world large-scale dataset~\cite{sun2020scalability}, which includes 374 training records and 76 test records in urban traffic scenarios. We train and evaluate the algorithms using the Waymo Open Dataset.

The data frames are collected by a top-mounted LiDAR sensor and are labeled by 3D bounding boxes with center position $(x, y, z)$ and size $(length, width, height)$ with unique track identifiers. We use 2D map representation $(x, y)$ in bird's-eye-view as input, together with unique track ids and type information. All pedestrians and vehicles within the scan range of the LiDAR sensor which is 75~m are considered. We pre-processed the Waymo Open Dataset by transforming the coordinates from local coordinates to global coordinates to reduce the influence of the ego-vehicle's movement.

We randomly separated the training dataset into 90\% training set (337 records) and 10\% validation set (37 records), and used the test set (76 records) for evaluation. We sampled the frames of the Waymo Open Dataset from 10 fps to 2.5 fps frequency to align the prediction with the previous works that used the ETH and UCY dataset at 2.5 fps. 
We cut the sequences into pieces with a fixed length when loading the data. The sequence length is set as the sum of the observation and prediction length. Here, we used 20 time-steps sequence lengths with 8 observation time-steps covering 3.2 seconds and 12 prediction time-steps covering 4.8 seconds. With skip step as one, we got 195,192 sequences for training, 36,946 for validation, and 52,484 for evaluation.
\subsection{Evaluation Metrics and Baselines}
\label{sec:Experiment baselines}
We use the following metrics to report the prediction error: 
\begin{itemize}
\item The Average Displacement Error (ADE): The average distance between ground truth and prediction trajectories overall predicted time-steps as defined in Eq.~\ref{eq_ade}.
\item The Final Displacement Error (FDE): The average distance between ground truth and prediction trajectories for the final predicted time-step as defined in Eq.~\ref{eq_fde}.
\end{itemize}
\begin{equation}
ADE=\frac{\sum_{i\in{n_p}}\sum_{t=T_{obs}+1}^{T_{pred}}{\| Y_t^i - \hat Y_t^i \|}_2}{n_p \times (T_{pred} - T_{obs})}
\label{eq_ade}
\end{equation}
\begin{equation}
FDE=\frac{\sum_{i\in{n_p}}{\| Y_t^i - \hat Y_t^i \|}_2}{n_p}, t = T_{pred}
\label{eq_fde}
\end{equation}
where $n_p$ is the number of pedestrians.

We apply the proposed PVI extractor on both sequential (LSTM-based) and non-sequential (Conv-based) models and get the following models:
\begin{itemize}
    \item SI-PVI-LSTM (ours): applies the PVI extractor on LSTM-based model. This model considers both social interaction (SI) and pedestrian-vehicle interaction (PVI).
    \item SI-PVI-Conv (ours): applies the PVI extractor on Conv-based model. This model considers both social interaction (SI) and pedestrian-vehicle interaction (PVI).
\end{itemize}

We compare the performance of our proposed models against the following baseline methods:
\begin{itemize}
    \item Linear Regression: A linear regression model of pedestrian motion over each dimension.
    \item LSTM: Na\"ive LSTM without the influence of the other pedestrians or vehicles.
    \item Social-LSTM: as proposed by Alahi et al.~\cite{alahi2016social}.
    \item Social-GAN: as proposed by Gupta et al.~\cite{gupta2018social}.
    \item Social-STGCNN: as proposed by Mohamed et al.~\cite{mohamed2020social}.
    \item Social-IWSTCNN: as proposed by Zhang et al.~\cite{zhang2021social}.
\end{itemize}

\subsection{Implementation Details}
The Nvidia GeForce RTX 2080 Ti GPU is used for the training and test process.
For the LSTM-based models, we trained the model with the Adam Optimizer. The training batch size was set to 16 for 200 epochs, and the learning rate was 1e-4. For the Conv-based models, we trained the model with the Stochastic Gradient Decent (SGD) with an initial learning rate of 0.01. The training batch size was set to 64 for 250 epochs. For the evaluation of all aforementioned methods, we generated 20 samples and used the closest sample to the ground truth to calculate the metrics.


\section{Results and Analysis}
\label{sec:ResultsAnalysis}
\subsection{Quantitative Results}
The methods listed in Sec.~\ref{sec:Experiment baselines} are trained and evaluated on the Waymo Open Dataset~\cite{sun2020scalability}. The ADE and FDE results (in meters) for 12 time-steps (ie., 4.8 seconds) prediction are shown in Table~\ref{table:results1}; lower results are better.
We compare the sequential and the non-sequential models separately. The bold font represents the best results.
The proposed SI-PVI-LSTM model outperforms the previous approaches Social-LSTM~\cite{alahi2016social} and Social-GAN~\cite{gupta2018social} on both ADE and FDE. The proposed SI-PVI-Conv model outperforms Social-STGCNN on both ADE and FDE, and outperforms Social-IWSTCNN on ADE.
The results demonstrate that the use of vehicle information improves the accuracy of pedestrian trajectory prediction.

\begin{table}[h]
\begin{center}
\caption{The ADE/FDE metrics for baseline methods compared to proposed methods on the Waymo Open Dataset.}
\label{table:results1}
\begin{tabular}{c|c|c|c|l}
\hline
\hline
\makecell{Prediction \\Method} & Model Name & ADE & FDE & \makecell[l]{Interactions Used \\ in the Model} \\
\hline
\hline
\multirow{4}{*}{Sequential} & LSTM & 0.392 & 0.844 &
\makecell[l]{No interactions.} \\
\cline{2-5}
& \makecell{Social-LSTM \\(2016)~\cite{alahi2016social}} & 0.402 & 0.840 & \makecell[l]{Social interaction.} \\
\cline{2-5}
& \makecell{Social-GAN \\(2018)~\cite{gupta2018social}} & 0.386 & 0.826 & \makecell[l]{Social interaction.} \\
\cline{2-5}
& \makecell{SI-PVI-LSTM \\(ours)} & \textbf{0.372} & \textbf{0.796} & \makecell[l]{Social interaction, \\Pedestrian-vehicle \\interaction.} \\
\hline
\multirow{4}{*}{\makecell{Non-\\sequential}} & Linear & 0.412 & 0.892 &\makecell[l]{No interactions.} \\
\cline{2-5}
& \makecell{Social-STGCNN\\(2020)~\cite{mohamed2020social}} & 0.334 & 0.550 & \makecell[l]{Social interaction.} \\
\cline{2-5}
& \makecell{Social-IWSTCNN\\ (2021)~\cite{zhang2021social}} & 0.329 & \textbf{0.540} & \makecell[l]{Social interaction.} \\
\cline{2-5}
& \makecell{SI-PVI-Conv \\(ours)} & \textbf{0.327} & 0.543 & \makecell[l]{Social interaction, \\Pedestrian-vehicle \\interaction.} \\
\hline
\hline
\end{tabular}
\end{center}
\end{table}

For the sequential models, the proposed SI-PVI-LSTM outperforms the LSTM, Social-LSTM, and the Social-GAN that do not include the pedestrian-vehicle interaction features. This shows that the pedestrian-vehicle interaction influences the pedestrian's behavior, and using this information can improve the prediction. Compared with Social-LSTM, using our proposed PVI extractor reduces the ADE and the FDE by 7.46\% and 5.24\%, respectively. Compared with the Social-GAN, the SI-PVI-LSTM gets better results without using the more complicated and hard-to-train GAN structure.

We notice that the Social-LSTM gets worse results than LSTM that only uses the individual information. This is consistent with the results on ETH and UCY datasets in Gupta et al.'s work~\cite{gupta2018social}.
We have improved the methods for extracting the social interaction by directly extracting the spatial features from the pedestrians' past positions in each frame, and using the spatial features to extract the interaction features.

For the non-sequential algorithms, LR performs worse than the other methods as expected, as it does not include any interaction information. Social-STGCNN and Social-IWSTCNN that consider the social interaction information reach more accurate results than the LR. The SI-PVI-Conv model uses the pedestrian-vehicle interaction information in addition to the social interaction, and achieves the best ADE result. Compared with Social-STGCNN, using our proposed PVI extractor reduces the ADE and FDE by 2.10\% and 1.27\%, respectively. 
Compared with Social-IWSTCNN, SI-PVI-Conv performs slightly better on ADE because of the additional vehicle information, but it does not improve the performance on FDE. There are two possible reasons for this. Firstly, the model only uses the vehicle information of the observation period, and as vehicles move much faster than pedestrians, the information may not be sufficient for a long-term prediction.
Secondly, in SI-PVI-Conv, we calculate the interaction with all vehicles in one frame and do not consider the orientation of the pedestrians. However, the vehicles behind the pedestrians may not influence the pedestrians as much as the other vehicles, and the included information may bring in extra noise.
Therefore, the FDE of SI-PVI-Conv does not improve yet still is comparative with the other two Conv-based models.

The Conv-based models get more accurate results than the LSTM-based models. This is because the Conv-based methods represent the moving states in a better way instead of using hidden-states of LSTMs. Moreover, recurrent predictions such as LSTM and LSTM-based GAN models will accumulate the error, while the Conv-based models do not have this drawback.

\begin{table}[htpb]
\begin{center}
\caption{Interaction and Influencing Factors of LSTM-based models}
\label{tab:sequential}
\begin{tabular}{c|cc|cccc}
\hline
\hline
\multirow{2}{*}{\makecell{Model Name}} & \multirow{2}{*}{ADE} & \multirow{2}{*}{FDE} & \multicolumn{4}{c}{Influencing Factors} \\ \cline{4-7}
 &  &  & \makecell{SI} & \makecell{PVI} & \makecell{RP} & \makecell{RV} \\
 \hline
 \hline
LSTM & 0.392 & 0.844 & F & F & F & F \\
\makecell{SI-LSTM} & 0.384 & 0.820 & T & F & T & F \\
\makecell{SI-PVI-LSTM} & \textbf{0.372} & \textbf{0.796} & T & T & T & F \\
\makecell{SI-LSTM (w. velocity) } & 0.390 & 0.823 & T & F & T & T \\
\makecell{SI-PVI-LSTM (w. velocity)} & 0.378 & 0.811 & T & T & T & T \\
\hline
\hline
\end{tabular}
\end{center}
\end{table}

The study of the influencing factors of pedestrian trajectory on LSTM-based models is shown in Table~\ref{tab:sequential}. We investigate the influence of the social interaction (SI), the pedestrian-vehicle interaction (PVI), and different inputs including the relative position (RP), and the relative velocity (RV).

For the LSTM-based models, with the improved social interaction (SI) extractor, the model SI-LSTM reduces the ADE and FDE significantly compared with LSTM, which shows that social interaction can influence the pedestrian's behavior.
This also provides evidence that the hidden states of LSTMs are unsuitable for calculating social interaction.
By including both the social interaction and pedestrian-vehicle interaction feature, the SI-PVI-LSTM improves the results and achieves the best performance. This shows that the pedestrian-vehicle interaction influences pedestrian behavior and we can improve the prediction accuracy by including vehicle information.

The comparison of Conv-based models is shown in Table~\ref{tab:non-sequential}. With only PVI information, the results are not improved compared with raw convolutional models that only use individual information.
The SI-PVI-Conv model gets the best results by including both the social interaction and the pedestrian-vehicle interaction, that shows both kinds of interactions are essential for prediction. 

\begin{table}[htpb]
\begin{center}
\caption{Interaction and Influencing Factors of Conv-based models}
\label{tab:non-sequential}
\begin{tabular}{c|cc|cccc}
\hline
\hline
\multirow{2}{*}{Model Name} & \multirow{2}{*}{ADE} & \multirow{2}{*}{FDE} & \multicolumn{4}{c}{Influencing Factors} \\ \cline{4-7}
 &  &  & \makecell{SI} & \makecell{PVI} & \makecell{RP} & \makecell{RV} \\
 \hline
 \hline
Conv & 0.334 & 0.571 & F & F & F & F \\
PVI-Conv & 0.346 & 0.586 & F & T & T & F \\
SI-PVI-Conv & \textbf{0.327} & \textbf{0.543} & T & T & T & F \\
\makecell{PVI-Conv (w. velocity)} & 0.341 & 0.556 & F & T & T & T \\
\makecell{SI-PVI-Conv (w. velocity)} & 0.339 & 0.562 & T & T & T & T \\ 
\hline
\hline
\end{tabular}
\end{center}
\end{table}

To investigate the inputs of extracting the interaction features, we compare the models with and without relative velocities between pedestrians and the other objects. The results are not improved by adding velocity information. One possible reason is that the pedestrians are very agile, and the velocity of the pedestrians can change all the time and may introduce noises into the network. 

We compared the inference speed of two competitive methods: Social-IWSTCNN and SI-PVI-Conv. The inference speed of SI-PVI-Conv is 3.39~ms per sequence, which is close to Social-IWSTCNN whose inference speed is 3.38~ms per sequence.
This shows that the computing of pedestrian-vehicle interaction features does not cost much inference time while can improve the accuracy.

\subsection{Qualitative Results}
\begin{figure*}[]
\begin{center}
\includegraphics[scale=0.25]{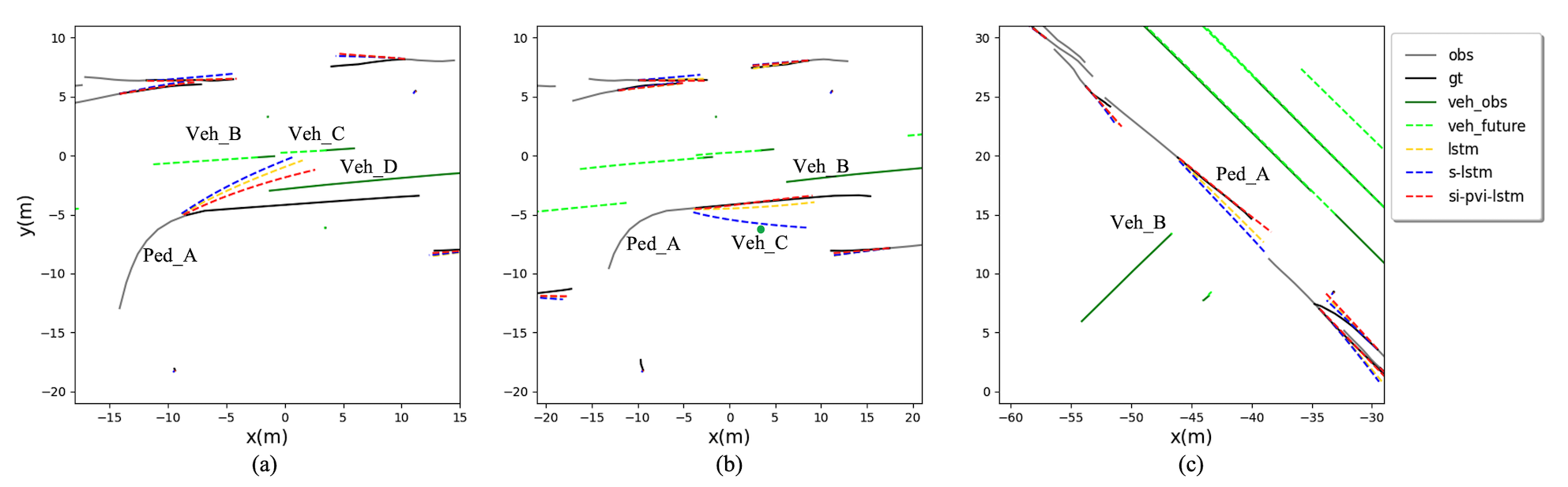}
\end{center}
  \caption{The comparison of prediction results of LSTM, Social-LSTM and SI-PVI-LSTM. (a) The scenario where pedestrian A is turning right, avoiding the moving vehicles B, C, and D.
  (b) The scenario where pedestrian A has turned right and keeps walking straight, avoiding moving vehicle B and parked vehicle C.
  (c) The scenario where pedestrian A is crossing the road, interacting with vehicle B that is slowing down and waiting.
  The legends: obs denotes for observed paths of pedestrians; gt refers to the ground truth of predicted trajectories of pedestrians. veh\_obs refers to the observed vehicle trajectories, and veh\_future stands for the future trajectories of vehicles during the prediction time. lstm refers to the LSTM model, s-lstm refers to Social-LSTM; si-pvi-lstm denotes our proposed Social Interaction and Pedestrian-Vehicle Interaction LSTM model.}
\label{fig:figure3}
\end{figure*}

To qualitatively analyze the performance of the PVI extractor, we take the LSTM-based model as an example, comparing the LSTM, Social-LSTM, and SI-PVI-LSTM models. The scenarios where the pedestrians interact with the vehicles are shown in Fig.~\ref{fig:figure3}.

In Fig.~\ref{fig:figure3} (a), the pedestrian A (denoted as Ped\_A) turns right. In the prediction of the SI-PVI-LSTM model, the pedestrian interacts with the moving vehicles B, C, and D (denoted as Veh\_B, Veh\_C, and Veh\_D in Fig.~\ref{fig:figure3} (a)), and keeps a distance to avoid collision with the vehicles. In the predictions of LSTM and Social-LSTM models, the trajectory of pedestrian A does not avoid vehicles B and C well, and may cause dangerous situations.
In Fig.~\ref{fig:figure3} (b), pedestrian A has turned right and keeps walking straight. In the prediction of the SI-PVI-LSTM model, pedestrian A keeps walking straight and avoids moving vehicle B and parked vehicle C keeping a safe distance. While in the result of Social-LSTM, the model does not use the vehicle information and the pedestrian does not interact with the vehicles, the predicted trajectory has a risk of colliding with the parked vehicle C. The SI-PVI-LSTM model provides a better prediction than the other two models that do not include the pedestrian-vehicle interaction.
In Fig.~\ref{fig:figure3} (c), pedestrian A is crossing the road, when vehicle B is slowing down and waiting. The prediction of the SI-PVI-LSTM model shows that pedestrian A tends to keep a safe distance from the vehicle, while in the results of LSTM and Social-LSTM models, pedestrian A does not.

In all these cases, the results show that the SI-PVI-LSTM model gains better performance, and can better capture the interaction between the pedestrian and vehicle than the Social-LSTM models. 
For the other pedestrians who do not have direct interaction with the vehicles, the results of Social-LSTM and SI-PVI-LSTM are similar. This indicates that the PVI extractor does not affect the feature learning of the social interaction between pedestrians.


\section{Conclusions and Future Work}
\label{sec:Conclusion}
In this paper, we present the Pedestrian-Vehicle Interaction (PVI) extractor to learn the pedestrian-vehicle interaction when predicting pedestrians' trajectories in urban traffic scenarios.
We extract the PVI features from vehicles' trajectories and their relative positions with the target pedestrians. In our framework, we include the spatial features of pedestrians, the social interaction features, and the PVI features.
To validate the effectiveness of the PVI extractor, we deploy the PVI extractor on both the sequence (LSTM-based) and non-sequence (convolutional-based) models.
We use the real-world traffic dataset to validate that our algorithm is suitable for urban traffic scenarios, and compare our work with sequential models, including LSTM, Social-LSTM, Social-GAN, and non-sequential models, including LR, Social-STGCNN, and Social-IWSTCNN. The quantitative and qualitative evaluations show that our algorithm is effective for urban traffic scenes and can capture the interaction between pedestrians and vehicles.
Future work may consider a new structure to better learn the interaction between pedestrians and vehicles. Besides, the environment scene feature and the appearance feature of the road users can also be included to improve the prediction. 

\bibliographystyle{IEEEtran}
\bibliography{IEEEexample}

\end{document}